\documentclass[10pt,twocolumn,letterpaper]{article}

\usepackage{wacv}
\usepackage{times}
\usepackage{epsfig}
\usepackage{graphicx}
\usepackage{amsmath}
\usepackage{amssymb}
\usepackage{subcaption}
\usepackage{multirow}
\usepackage[table,xcdraw]{xcolor}
\usepackage{adjustbox}
\usepackage{hhline}
\usepackage{enumitem}

\wacvfinalcopy 

\def\wacvPaperID{242} 

\ifwacvfinal\fi
\begin{document}

\title{Coupled Generative Adversarial Network for Continuous Fine-grained Action Segmentation}

\author{Harshala Gammulle \hspace{.4cm} Tharindu Fernando  \hspace{.4cm}  Simon Denman \hspace{.4cm} Sridha Sridharan \hspace{.4cm} Clinton Fookes\\
Image and Video Research Laboratory, SAIVT, Queensland University of Technology, Australia.\\
{\tt\small  \{pranali.gammule, t.warnakulasuriya, s.denman, s.sridharan, c.fookes\}@qut.edu.au}
}


\maketitle
\ifwacvfinal\thispagestyle{empty}\fi

\begin{abstract}
  We propose a novel conditional GAN (cGAN) model for continuous fine-grained human action segmentation, that utilises multi-modal data and learned scene context information. The proposed approach utilises two GANs: termed Action GAN and Auxiliary GAN, where the Action GAN is trained to operate over the current RGB frame while the Auxiliary GAN utilises supplementary information such as depth or optical flow. The goal of both GANs is to generate similar `action codes', a vector representation of the current action. To facilitate this process a context extractor that incorporates data and recent outputs from both modes is used to extract context information to aid recognition. The result is a recurrent GAN architecture which learns a task specific loss function from multiple feature modalities. Extensive evaluations on variants of the proposed model to show the importance of utilising different information streams such as context and auxiliary information in the proposed network; and show that our model is capable of outperforming state-of-the-art methods for three widely used datasets: 50 Salads, MERL Shopping and Georgia Tech Egocentric Activities, comprising both static and dynamic camera settings.\footnote{This research was supported by the Australian Research Council's Linkage Project LP140100282 ``Improving Productivity and Efficiency of Australian Airports''}
\end{abstract}

\section{Introduction}
\vspace{-2mm}

In this paper, we propose a coupled Generative Adversarial Network approach for continuous action segmentation. Action segmentation is performed as in \cite{leaCVPR}, by predicting the action occurring at every video frame, considering all action classes together with the `Background' (action transitions). We treat the action segmentation process as a generative problem where the generator learns to generate an action code which represents a coded distribution of the action categories present in the current frame.

\begin{figure}[!h]
        \centering
        	\includegraphics[width=0.85\linewidth]{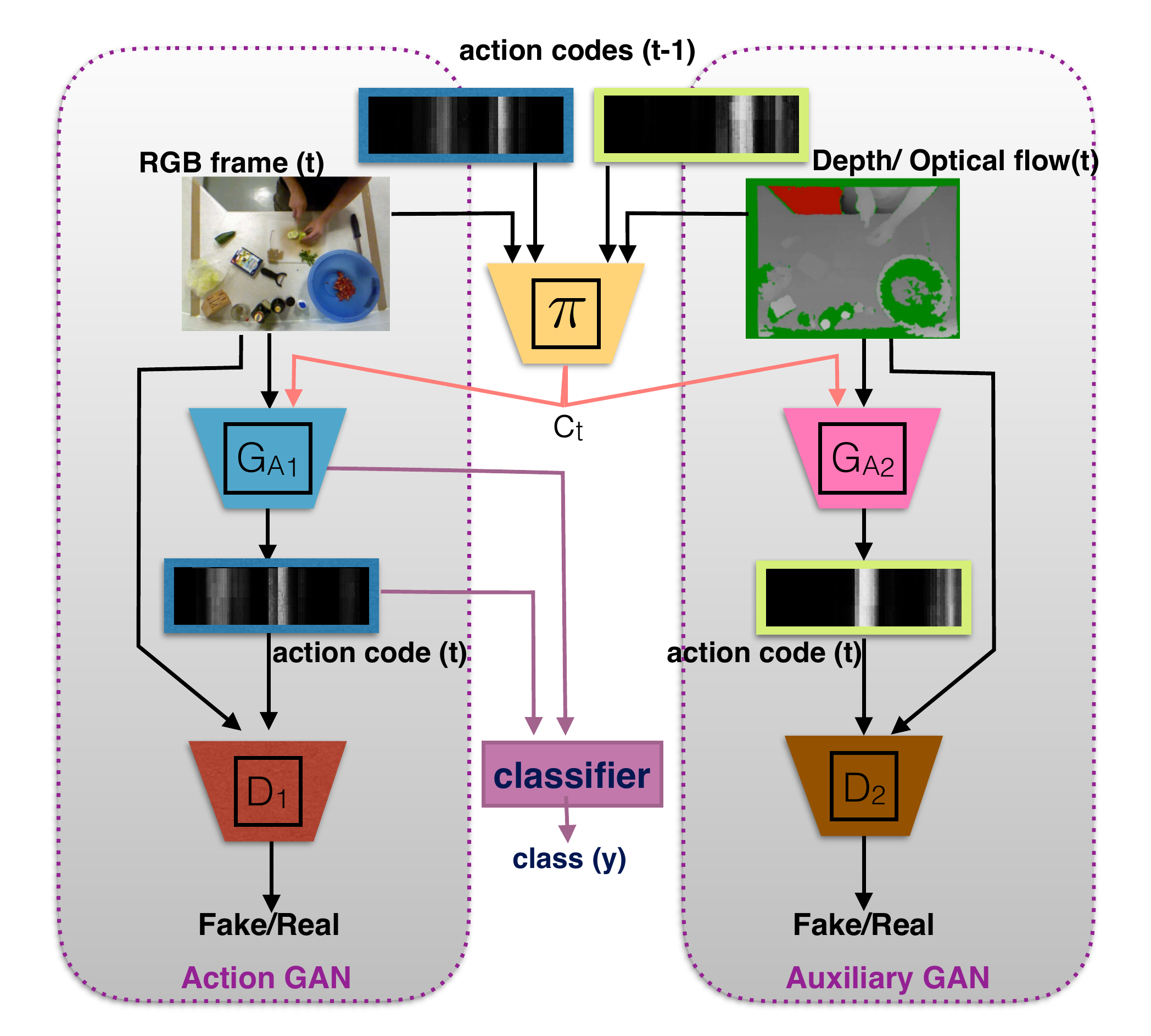}
	\caption{The proposed \textit{Coupled Action GAN}: The action GAN (left) takes the RGB frame as input and the auxiliary GAN (right) takes supplementary information (depth or optical flow). Both networks are aided by context information from the context extractor ($\pi$), which recieves the previously generated action codes from both action generators, and the current RGB and auxiliary frames. Additionally, a classifier is used with the action GAN generator to reinforce it's performance.}
	\label{fig:system_diagram}
	\vspace{-3mm}
\end{figure}

Most recent works in continuous action segmentation use deep neural networks to process spatio-temporal information, and as such do not require feature engineering or suffer from the inherent deficiencies of hand crafted feature-based methods. Even though deep network based approaches learn to minimise a loss function automatically, manual effort is required to design an effective loss to obtain optimal results \cite{Isola_CVPR2017}. One approach to overcome this is generative adversarial networks \cite{GAN2014}, which automatically learn a loss function while trying to generate an output indistinguishable from real data. This is achieved using two main components, a `generator' and a `discriminator', where the ultimate goal of the `generator' is to create fake outputs that are difficult for the `discriminator' to distinguish from real ones. Recent research has shown GANs to be effective for diverse tasks including future frame prediction \cite{mathieu2015}, image synthesis~\cite{text2imageGAN}, domain adaptation~\cite{kiasari2018coupled,liu2016coupled}, inpainting \cite{yoo2016,inpainting}, and visual saliency prediction \cite{pan2017salgan,fernandoTask}. However, as yet there are no GAN based methods for detection and recognition problems where the processing of a temporal data stream is required.

We address this by proposing a conditional GAN based approach for continuous action recognition (see system diagram in Figure \ref{fig:system_diagram}). The proposed method has three main components: the action GAN, the auxiliary GAN and the context extractor. The action GAN takes the current RGB input frame while the auxiliary GAN uses corresponding supplementary information (depth or optical flow images). Both networks are supported by the context information produced by the context extractor, which combines data from both modes to extract additional salient details from the environment to aid recognition. Here, the importance of using environmental context is that it captures information related to the environment and recent actions, which may include details such as items present and interactions between them and the subject.

%

Both the action and auxiliary GANs aim to generate a similar dense vector representations, an `action code', that represents the current action in the video. The discriminator of each GAN then uses the generated action code and the frame (RGB or auxiliary) to classify whether the action code is generated (`fake') or ground truth (`real'). We reinforce the action code generation process with a classifier loss, forcing generated codes to be informative for the final action classification task that we are interested in.  

We aim to capture low-level information in the input frame through the action generator, and high level through the context-extractor through which we also achieve temporal modelling. We treat the input sequence-wise, and synthesised action codes for the previous frame are an input to the context extractor, hence the model is recurrent. In contrast to frameworks which predict start/end frames for events and then recognise actions for segments, we predict the class of each frame, including the background class, and thus temporally segment the input video. As such, our evaluation uses unsegmented continuous action datasets, but could also be applied to pre-segmented datasets. 

The main contributions of this work are:

\begin{enumerate}[label=\alph*]
\item We propose a novel recurrent GAN architecture which combines multiple feature modalities and scene context information for action segmentation. 
\item We introduce a simple yet effective method to augment the performance of the generator by using classification loss as a reinforcement signal.
\item We provide experiments on three challenging real world databases where in all cases, the proposed method outperforms state-of-the-art approaches.   
\item We perform an ablation study on different architectural variants of the proposed model to identify key contributions of different input streams, and show how the components of the proposed approach work together to generate state of the art results.
\end{enumerate}

\section{Related work}
\vspace{-2mm}

The task of recognising actions in video has been a popular research area in computer vision. There exist two major lines of work which address the problem of action recognition. The first category of methods are discrete in nature and operate on images \cite{imageBased1} or pre-segmented videos that contain only a single action \cite{twoStream,gammulle2017two,actiontubes,3dcnn}. Even though such approaches can achieve high accuracy, they represent an oversimplification of the task (as they operate on pre-segmented videos containing one action per video), making them unsuited for real world problems such as detecting threats from a continuous CCTV feed. This limitation is the motivation for researchers to focus on continuous action datasets \cite{cooking1,merlshopping,50salads,breakfast} that contain fine-grain actions.

Numerous action detection and segmentation methods have been introduced over the past few years to address the challenge of continuous action recognition. Some methods are based on extracting hand-crafted features to model fine-grain actions. In \cite{rohrbach2016,cooking1} Rohrbach et al. introduced methods that utilise pose related and dense trajectory features extracted from HOG, optical flow \cite{laptev2008} and motion boundary histograms (MBH) around densely sampled points in the MPII cooking activity dataset \cite{cooking1}. Kuehne et al. \cite{kuehne2016end,breakfast} model actions using a Hidden Markov model on dense trajectory features. In \cite{50salads}, Stein et al. introduced a method for fusing accelerometers and computer vision for the purpose of recognising actions in the 50 salads database using classifiers such as naive Bayes \cite{naivebay} and random decision forests \cite{decForest}. Several other works introduced methods based on object centric feature extraction \cite{Ni2016,ramirez2014}. However these object based methods are limited to actions involving object interactions and are specifically tailored for each dataset, and thus are not easily transferred to a different domain. 

Aside from using handcrafted features, approaches have been introduced using deep networks. Singh et al. \cite{merlshopping} introduced a multi-stream bi-directional recurrent neural network utilising both spatial and temporal information; while Lea et al. \cite{lea2016seg} incorporates a spatio-temporal CNN with a constrained segmental model. In \cite{leaCVPR}, the authors have introduced temporal convolutional networks (TCN) for fine grained action detection and segmentation.  

However these methods, whether using handcrafted features or deep neural networks, still need manual human effort to perform well; requiring either effort to design feature representations, or to effective loss functions. As a result of this, effort is being channeled to new directions such as Generative Adversarial Networks (GAN). GANs are capable of learning outputs that are difficult to discriminate from real examples, and learn a mapping from input to output while learning a loss function to train the mapping. GANs have been applied to solve different computer vision based problems such as future frame \cite{mathieu2015} or state predictions \cite{zhou2016}, product photo generation \cite{yoo2016}, and inpainting \cite{inpainting}. As these methods are image generation methods some studies have sought to add temporal information \cite{seqgan,kannan2017,tgan}, to extend GANs to classification based approaches \cite{denton2016} and synthesise images from a text description \cite{text2imageGAN}. The works of \cite{seqgan,kannan2017,tgan} investigated modelling sequences with a GAN where the generator learns to create a sequence and the discriminator outputs a classification for the entire sequence. However for action segmentation we argue a classification is needed for each frame, rather than for the entire sequence, and hence we cannot directly utilise the above methods. 

Several attempts have been made to couple multiple GANs for domain adaptation. Specifically \cite{kiasari2018coupled} and \cite{liu2016coupled} proposed coupled GAN architectures for generating images in different domains with a joint random vector input. However we are considering multi-modal input streams of the same action representation, and are modelling videos as opposed to a single image. The above stated coupling mechanisms do not utilise multiple input streams, and consider a common random vector input. The coupling is achieved through weight sharing between the 2 generators which is unsuitable in our case as we have diverse inputs (i.e RGB, depth or optical flow) and although they represent the same action, the vastly different modalities necessitate separate generators. Here we propose to perform coupling through a context extractor network,  which captures salient information from both modes. Through the context extractor we also obtain a recurrent architecture by using the action code generated at the previous time step as an input in the current time step. Such temporal modelling is essential as we are considering video sequences rather than single images.

\section{Methodology} 
\vspace{-2mm}
\label{metho}

We utilise the conditional GAN \cite{condGAN} (an extension of the GAN), which learns a mapping from the observed image $x_{t}$ at time $t$ and a random noise vector $z_{t}$ to $y_{t}$ \cite{Isola_CVPR2017}. These GANs have two main components: a Generator G, which creates outputs that it aims to make difficult to distinguish from real data by an adversarially trained discriminator, D, that tries to detect the fake outputs generated by G. 

We introduce a conditional GAN based model, \textit{coupled action GAN}, for continuous fine-grained action segmentation, which couples spatial and temporal information to improve performance. In Section \ref{ac_code}, we describe the action code format that we train the GAN to create; in Section \ref{obj} we explain the objectives behind our models and in Section \ref{coupling_info} we explain the coupled network behaviour.  

\subsection{Action codes}
\vspace{-2mm}
\label{ac_code}

The generators output an action code representing the action category of each frame. The generator maps dense pixel information to this action code. Hence having a one hot vector is not optimal. Therefore we scale it to a range from 0 to 255, $y_{t}  \epsilon  {\rm I\!R}^{1 \times{k}}$, where k is the number of action classes in the dataset. Using an integer encoding gives more freedom for the action generator and discriminator to represent each action code as a dense vector representation. The action code can be seen as an intermediate representation of the input frame which is more distinguishable during the classification process, and prior works have shown the value of such representations with GANs~\cite{rebuttal_song2017}.
%

\subsection{Objectives}
\vspace{-2mm}
\label{obj}

\begin{figure}[!htb]
    \centering
    \begin{subfigure}{.3\columnwidth}
        \includegraphics[width=.95\columnwidth]{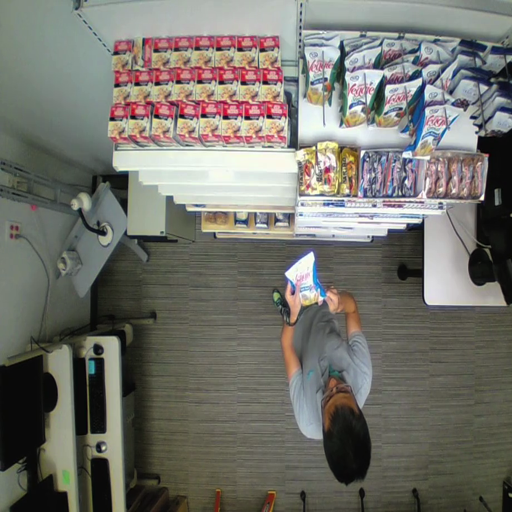} %
       
    \end{subfigure}
    \begin{subfigure}{.3\columnwidth}
        \includegraphics[width=.95\columnwidth]{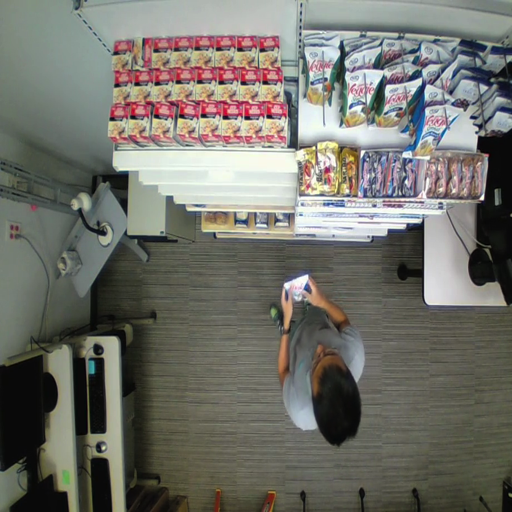}
       
     \end{subfigure}
 \caption{ Visually similar frames from different action classes: $background$ (left) and $Inspect\_Product$ (right)}
	\vspace{-3mm}
  \label{fig:similar}
\end{figure}

Conditional GANs learn both a mapping from input to output, and a loss to train this mapping. Therefore, they are suitable for problems that require varying loss formulations. The objective for the conditional GAN is defined as,

\begin{equation}
\begin{split}
L_{cGAN}(G,D)=\mathbb{E}[\log D(x_t,y_t)]+ \\ \mathbb{E}[\log(1-D(x,G(x_t,z_t))],
\end{split}
\label{eq:3}
\end{equation}  

Many visual recognition tasks critically rely on context \cite{context}. This is especially the case in situations where there are actions that are visually similar (as in Figure \ref{fig:similar}), but belong to different classes \cite{Shapovalova2011,santofimia2014,vrigkas2015,Arunothayam2012}. Therefore, we employ an additional context extractor to enhance the model.

As shown in Figure \ref{fig:system_diagram}, the inputs for the context extractor, $\pi$, are the current frames (RGB with the auxiliary data) and the previous action distribution codes obtained by the action generators. Note that the two GANs are formulated and trained in the same manner. The only differences between the two are the input data, and that for the Action GAN (trained with RGB data) a secondary classifier is trained from intermediate outputs to improve learning.

After coupling the generator $G_{A_1}$ with the context extractor, the adversarial loss for the GANs ($V_{G_{A_1},D_1}$) can be defined as follows,

\begin{equation}
\begin{split}
V_{G_{A_1},D_1}=\mathbb{E}[\log(D_1(x_{t},y_{t}))]+ \\ \mathbb{E}[\log(1-D_1(x_{t}, G_{A_{1}}(x_{t},z_{t},c_{t})))],
\end{split}
\label{eq:4}
\end{equation} 
where $c_{t}$ is the output of $\pi$ at time instance t.

Both action generators are trained to output an action code which provides a distribution over all possible action classes. However, our purpose is to classify the frames into discrete action classes. Therefore, we define a classifier that extracts a number of features from intermediate layers of the action generator model ($G_{A_{1}}$, i.e. the generator that uses RGB images). Let $[\theta_{0},\theta_{1},..., \theta_{n}]$ be the features extracted from $n$ layers where $\theta_{i}=f_{i}(G_{A_1}(x_{t},c_{t}))$ is a function that extracts features from the $i^{th}$ layer of the action generator. Then the loss of the multi class classification function $f_{c}$ is,

\begin{equation}
\begin{split}
\mathcal{L}_{c}=\mathbb{E}[\log f_{c}(f_{0}(G_{A_1}(x_{t},z_{t},c_{t}))),...,\\ f_{n}(G_{A_1}(x_{t},z_{t},c_{t})))].
\end{split}
\label{eq:5}
\end{equation} 

 $\mathcal{L}_{c}$ is then used to reinforce the GAN objective with classification error,

\begin{equation}
V^*_{G_{A_{1}},D_1}=V_{G_{A_1},D_1}-\lambda_{1} \mathcal{L}_{c}
\label{eq:6}
\end{equation} 

We use a softmax classifier \cite{bishop2006pattern,zeng2014relation} to classify the action class. As all functions are differentiable we train the entire model end-to-end using back-propagation. 

\subsection{Coupling multi-model information}
\vspace{-2mm}
\label{coupling_info}

The use of multi-model information benefits a recognition approach as the modes can represent different aspects of the actions \cite{twoStream}. We couple multiple action generators, $G_{A_{1}}$ and $G_{A_{2}}$, as shown in Figure \ref{fig:system_diagram}. Here the second GAN is coupled as an auxiliary network, which takes supplementary information. This supplementary information may vary across datasets; for instance we use depth information for the 50 salads dataset \cite{50salads} and optical flow for MERL shopping \cite{merlshopping} and Georgia Tech Egocentric activity \cite{GTEA_1} datasets. Both GANs aim to generate realistic action codes to fool their respective discriminators using their differing inputs, and the coupled adversarial loss can be defined as, 

\begin{equation}
V_{G_{A_1},G_{A_2},D_{1},D_{2}}= V^*_{G_{A_{1}},D_{1}}+V_{G_{A_{2}},D_{2}},
\label{eq:7}
\end{equation} 

\begin{equation}
\begin{split}
V_{G_{A_1},G_{A_2},D_{1},D_{2}}= \mathbb{E}[\log(D_{1}(x_{t},y_{t}))]+ \\ \mathbb{E}[\log(1-D_{1}(x_{t}, G_{A_{1}}(x_{t},z_{t},c_{t})))]+\\ \mathbb{E}[\log(D_{2}(x'_{t},y'_{t}))]+ \\ \mathbb{E}[\log(1-D_{2}(x'_{t}, G_{A_{2}}(x'_{t},z_{t},c_{t})))]-\lambda_{1} \mathcal{L}_{c} ,
\end{split}
\label{eq:8}
\end{equation} 

where $x'_{t}$ is the input to the auxiliary network at time instance t, $y'_{t}$ is the generated action code from $G_{A_{2}}$ and $D_1$ and $D_2$ are the respective discriminators for the RGB and auxiliary streams. 

$G_{A_{2}}$ (auxiliary GAN) differs from $G_{A_{1}}$ in that the multi-class classifier (Equation \ref{eq:6}) is not applied, and so Equation \ref{eq:4} is used as the adversarial loss for $G_{A_{2}}$. As both $G_{A_{1}}$ and $G_{A_{2}}$ are observing different modalities of the same action at the same time step, and ideally the action codes generated by both $G_{A_{1}}$ and $G_{A_{2}}$ should be similar. Hence, feeding the classifier with both modalities is redundant. Therefore, we use features from $G_{A_{1}}$ only and reinforce only the objective of $G_{A_{1}}$ with classification loss.    

Network coupling occurs through the loss function. Both GANs try to generate similar codes to represent the current action. The primary stream for segmentation is action GAN and it's outputs are used for the final classification result. However both GANs are used as inputs to the context extractor, and so both influence the final decision.

If we try to reinforce the objective of the auxiliary GAN with classification loss the gradients may be smaller (hence ineffective) as $G_{A_2}$ and the classifier are not directly connected. However through the proposed context extractor we force $G_{A2}$ to generate features that are informative for $G_{A_1}$ for classification. This stabilises the training process (see Fig. 2 in supplementary material showing the convergence). If we use 2 classifiers, one each for $G_{A_1}$ and $G_{A_2}$, the objective of $G_{A_2}$ and the classification objective will attempt to force the representation of $G_{A_2}$ to be optimal classification performance. Hence rather than allowing $G_{A2}$ to learn features complementary to $G_{A_1}$, this would seek to generate a representation for classifying actions using only the auxiliary stream, discarding it's relationship with the context extractor and $G_{A_1}$. Finally, having two classifiers is also redundant as we only seek a single classification output.

\section{Network Architecture}
\vspace{-2mm}

%
%

The input RGB frame is of size of $224\times{224}\times{3}$. We also reshape optical flow and depth maps to sizes $224\times{224}\times{2}$ and $224\times{224}\times{1}$ respectively. The networks used in our model contain modules of the form : 2D convolution, followed by a batch normalisation, and a ReLU activation which we denote convolutional\_BatchNorm\_ReLu as in \cite{ioffe2015}. All convolutions are $4 \times{4}$ filters applied with a stride of two such that the output is down sampled by a factor of 2. Specific details for each of the networks (context extractor, action generator, and discriminator) are outlined in Section \ref{c_g}, \ref{a_g} and \ref{d_g} respectively. In Section \ref{opt} we present other details relevant to model training.

\subsection{Context Extractor}
\label{c_g}
\vspace{-2mm}

The context extractor receives two visual inputs: the RGB and auxiliary input frame. Each input is passed through a network with five convolutional\_BatchNorm\_ReLu layers containing 64, 128, 256, 512 and 512 kernels respectively. The output of the fifth layer (C512) of each stream is flattered and concatenated with the previous action distribution generated by $G_{A_{1}}$ and $G_{A_{2}}$ for RGB and auxiliary streams respectively. Finally we pass the embedding through a fully connected layer of size 256, generating a context embedding of size 256. Concatenation is done after encoding the input image as prior concatenation can lead to information loss of the previous action code. When using a single input (see Section \ref{ablation}), we omit one of the convolutional\_BatchNorm\_ReLu chains, and use the single available context vector. All other network parameters and layer sizes stay the same.

\subsection{Action Generators}
\label{a_g}
\vspace{-2mm}
As the generator network for both $G_{A_{1}}$ and $G_{A_{2}}$ we utilise the encoder architecture introduced in \cite{Isola_CVPR2017} as it is effective for visual information embedding. In the network, the input (i.e. RGB frame for $G_{A_{1}}$ or auxiliary input for $G_{A_{2}}$), is passed through eight convolutional\_BatchNorm\_ReLu layers. The flattened data is concatenated with the context extractor output before passing through three final fully connected layers. The output of the action generator is of size $k$ units where $k$ is the number of action classes in the dataset. $G_{A_{1}}$ and $G_{A_{2}}$ consist of convolutional\_BatchNorm\_ReLu layers containing 64, 128, 256, 512, 512, 512, 512 and 512 kernels followed by fully connected layers, with ReLu activation, of sizes 256, 128 and k where k is the number of action classes in the dataset. 

\subsection{Discriminators}
\label{d_g}
\vspace{-2mm}

The network either receives the RGB or auxiliary frame as input, depending on which generator it's paired with, along side the action code. We pass the image input through the two convolutional\_BatchNorm\_ReLu layers before concatenating it with the dense action code representation. The network then outputs whether the input action code is real or fake. A shallow discriminator architecture is used following~\cite{roth2017NIPS,neyshabur2017}, where the authors found it hard to train deep GANs due to unstable gradients. The discriminator architecture consists of two convolutional\_BatchNorm\_ReLu layers with 64 and 128 kernels followed by a fully connected layer with size 1 and softmax activation.

\subsection{Network Training}
\label{opt}
\vspace{-2mm}
We follow the training method of \cite{Isola_CVPR2017} and alternate between gradient decent passes for the discriminators and the generators, using minibatch SGD (32 examples per minibatch) and the Adam optimiser \cite{adam2015} with an initial learning rate of 0.1 for 25 epochs, and 0.01 for the next 75 epochs. No guidance is provided for the context extractor, and it jointly back propagates with the generators, learning to output informative embeddings. The classifier extracts features from the $8^{th}$, $10^{th}$ and $12^{th}$ layers of $G_{A_{1}}$, and concatenates them before parsing them to the softmax classifier.

\section{Evaluation and Discussion}
\vspace{-2mm}
In this Section we present the datasets (Section \ref{ds}) and metrics used (Section \ref{met}); the performance of the proposed approach compared to state-of-the-art (Section \ref{results}). Section \ref{ablation} presents an ablation study demonstrating the value of multiple inputs and augmentations such as context. 
    
\subsection{Datasets}
\vspace{-2mm}
\label{ds}

\textbf{The University of Dundee 50 Salads Dataset \cite{50salads}} contains 50 video sequences of 25 users, each making a salad in two different videos. Each sequence is 5-10 minutes long and obtained from a static RGBD camera pointed at the user. It is a multi-modal dataset including depth and accelerometer data alongside time-synchronised videos. However, we utilise only the video data for evaluation purposes. All seventeen mid-level action classes are used together with the background class. 
 
\textbf{ The MERL Shopping Dataset \cite{merlshopping}} contains 96 videos (32 subjects, 3 videos per subject), each two minutes long from a static overhead HD camera showing people shopping from grocery-store shelving units. Videos are composed of five action classes other than the background class.   

\textbf{The Georgia Tech Egocentric Activities (GTEA) Dataset \cite{GTEA_1}} is composed of videos recorded from a head mounted camera and includes four subjects performing seven different daily activities. This dataset comprises a dynamic, egocentric camera setting which is significantly different from the static top view of the previous 2 datasets. We utilise 11 action classes defined in \cite{gtea_evalpaper} including the background class. The evaluation is done as described in \cite{leaCVPR}.  

As the supplementary inputs for the auxiliary network, we feed the available depth maps for the 50 Salads Dataset and optical flow images for MERL Shopping and GTEA.
\subsection{Metrics}
\vspace{-2mm}
\label{met}
To comprehensively evaluate the proposed approach we use segmentation and frame wise accuracy metrics. Frame wise metrics are widely used~\cite{lea2016seg,leaCVPR,50salads,breakfast}, however as noted by \cite{leaCVPR} models having similar frame wise accuracies can show large dissimilarities when visualising their performance due to different segmentation behaviour. Therefore, using only frame wise metrics is insufficient to fully describe performance. Considering this, we also use the segmentation metrics: mean average precision with midpoint hit criterion (mAP@mid) \cite{merlshopping,rohrbach2016}, Segmental F1 score (F1@k) \cite{leaCVPR} and segmental edit score (edit) \cite{lea2016}.

\subsection{Results}
\vspace{-2mm}
\label{results}

For all datasets, we consider the Temporal Convolutional Networks (TCN) action detection and segmentation approach of \cite{leaCVPR} as a baseline. In \cite{leaCVPR}, they introduce two architectures: encoder-decoder TCN (ED-TCN) and dilated TCN; where ED-TCN uses pooling and up sampling to capture long range temporal patterns while dilated TCN uses dilated convolutions. 

For the 50 salads dataset we also consider the state-of-the-art methods introduced in \cite{lea2016seg,IDT_LM}. Richard et al. \cite{IDT_LM} includes statistical length and language modelling to represent temporal and contextual structure, and performs detection and classification jointly. In \cite{lea2016seg} Lea et al. introduced a fine-grain action segmentation method using spatio-temporal CNNs able to capture information such as object states, their relationships and their changes over time. We compare to their two proposed models: Spatial CNN and ST-CNN (see Table \ref{tab:tab_new}). The TDRN \cite{TDRN2018} model could be seen as an extension of ED-TCN where the authors replace the temporal convolution layers of the ED-TCN model using deformable temporal convolutions, allowing the model to capture fine-scale temporal details, in contrast to the fixed temporal receptive size of ED-TCN.

For the MERL shopping dataset, we compare the proposed approach to the methods introduced by Singh et al. \cite{merlshopping}. We use the results provided by \cite{leaCVPR}, as \cite{leaCVPR} re-evaluated Singh et al's \cite{merlshopping} models using segmentation and frame wise metrics. The `MSN Det' results are the sparse set of action detections, while the results for `MSN Seg' are a set of dense (per frame) action segmentations. 

For the Georgia Tech egocentric activities dataset, comparison is made to the results provided in \cite{singh2016Egocentric}, who have proposed a CNN network termed `Ego ConvNet' with two streams (a spatial and temporal stream) as introduced in \cite{twoStream}. The remaining systems presented are based on the results obtained through the models Spatial CNN, ST-CNN, Dilated TCN, Bi-LSTM, ED-TCN and TDRN.      

When considering the results presented in Table \ref{tab:tab_new}, we observe similar frame wise classification accuracies for Spatial CNN \cite{lea2016seg}, Dilated TCN \cite{leaCVPR}, ST-CNN \cite{lea2016seg}, Bi-LSTM \cite{leaCVPR}, ED-TCN \cite{leaCVPR} TricorNet and  TDRN \cite{TDRN2018}. But significant variations between F1 scores are seen due to over segmentation by the different approaches.

The proposed  GAN framework is capable of learning the hierarchical structure of the input frames along with the generated action codes, enabling improved classification of actions. Furthermore, in contrast to the Bi-LSTM, ED-TCN, TricorNet and TDRN models, we model the temporal context as a separate information stream. We believe this enables the proposed model to oversee the evolution of sub-actions and the relationships between them more effectively. This emphasises the importance of capturing auxiliary information available in the dataset and properly localising the present context through the multi-model information streams. These additional data cues along with the ability of GANs to learn a task specific loss allow the proposed model to outperform the state-of-the-arts in both action segmentation and frame wise classification. 

\begin{table}[h!]
\centering
\resizebox{1\linewidth}{!}{
\begin{tabular}{|p{2cm}|c|c|c|c|c|}
\hline
Dataset                                              & Approach                                   & F1@\{10,25,50\}                                   & edit                                  & mAP@mid                               & Acc                              \\ \hline
                                                     & Spatial CNN \cite{lea2016seg}                                & 32.3, 27.1, 18.9                                  & 24.8                                  & NA                                    & 54.9                                  \\ \cline{2-6} 
                                                     & IDT+LM                                     & 44.4, 38.9, 27.8                                  & 45.8                                  & NA                                    & 48.7                                  \\ \cline{2-6} 
                                                     & Dilated TCN                                & 52.2, 47.6, 37.4                                  & 43.1                                  & NA                                    & 59.3                                  \\ \cline{2-6} 
                                                     & ST-CNN                                     & 55.9, 49.6, 37.1                                  & 45.9                                  & NA                                    & 59.4                                  \\ \cline{2-6} 
                                                     & Bi-LSTM                                    & 62.6, 58.3, 47.0                                  & 55.6                                  & NA                                    & 55.7                                  \\ \cline{2-6} 
                                                     & ED-TCN                                     & 68.0, 63.9, 52.6                                  & 59.8                                  & NA                                    & 64.7                                  \\ \cline{2-6}
                                                     & TricorNet \cite{tricornet}             & 70.1, 67.2, 56.6                                  & 62.8                                  & NA                                    & 67.5   				\\ \cline{2-6}
                                                     & TDRN \cite{TDRN2018}             & 72.9, 68.5, 57.2                                  & 66.0                                  & NA                                    & 68.1                                  \\ \cline{2-6} 
\multirow{-7}{2cm}{50 Salads \cite{50salads}}                          & \cellcolor[HTML]{C0C0C0} Proposed & \cellcolor[HTML]{C0C0C0}\textbf{80.1, 78.7, 71.1} & \cellcolor[HTML]{C0C0C0}\textbf{76.9} & \cellcolor[HTML]{C0C0C0}\textbf{79.1}     & \cellcolor[HTML]{C0C0C0}\textbf{74.5} \\ \hline \hline
                                                     & MSN Det                                    & 46.4, 42.6, 25.6                                  & NA                                    & 81.9                                  & 64.6                                  \\ \cline{2-6} 
                                                     & MSN Seg                                    & 80.0, 78.3, 65.4                                  & NA                                    & 69.8                                  & 76.3                                  \\ \cline{2-6} 
                                                     & Dilated TCN                                & 79.9, 78.0, 67.5                                  & NA                                    & 75.6                                  & 76.4                                  \\ \cline{2-6} 
                                                     & ED- TCN                                    & 86.7, 85.1, 72.9                                  & NA                                    & 74.4                                  & 79.0                                  \\ \cline{2-6} 
\multirow{-5}{2cm}{MERL Shopping \cite{merlshopping}}                      & \cellcolor[HTML]{C0C0C0} Proposed & \cellcolor[HTML]{C0C0C0}\textbf{92.8, 91.7, 86.2} & \cellcolor[HTML]{C0C0C0}\textbf{89.1}     & \cellcolor[HTML]{C0C0C0}\textbf{89.8} & \cellcolor[HTML]{C0C0C0}\textbf{92.6} \\ \hline \hline
                                                     & EgoNet+TDD                                 & NA                                                & NA                                    & NA                                    & 64.4                                  \\ \cline{2-6} 
                                                     & Spatial CNN                                & 41.8, 36.0, 25.1                                  & NA                                    & NA                                    & 54.1                                  \\ \cline{2-6} 
                                                     & ST-CNN                                     & 58.7, 54.4, 41.9                                  & NA                                    & NA                                    & 60.6                                  \\ \cline{2-6} 
                                                     & Dilated TCN                                & 58.8, 52.2, 42.2                                  & NA                                    & NA                                    & 58.3                                  \\ \cline{2-6} 
                                                     & Bi-LSTM                                    & 66.5, 59.0, 43.6                                  & NA                                    & NA                                    & 58.3                                  \\ \cline{2-6} 
                                                     & ED- TCN                                    & 72.2, 69.3, 56.0                                  & NA                                    & NA                                    & 64.0                                  \\ \cline{2-6} 
                                                      & TricorNet \cite{tricornet}             & 76.0, 71.1, 59.2                                  & NA                                    & NA                                    & 64.8  				\\ \cline{2-6}
                                                      & TDRN \cite{TDRN2018}            & 79.2, 74.4, 62.7                                  & 74.1                                   & NA                                    & 70.1                                  \\ \cline{2-6}
\multirow{-7}{2cm}{GTEA dataset \cite{GTEA_1}} & \cellcolor[HTML]{C0C0C0}Proposed & \cellcolor[HTML]{C0C0C0}\textbf{80.1, 77.9, 69.1} & \cellcolor[HTML]{C0C0C0}\textbf{72.8}     & \cellcolor[HTML]{C0C0C0}\textbf{78.1}     & \cellcolor[HTML]{C0C0C0}\textbf{78.5} \\ \hline
\end{tabular}}
\caption{Action segmentation results for 50 Salads, MERL Shopping and Georgia Tech Egocentric Activities datasets : F1@k is the segmental F1 score, edit is the segmental edit score metric (see \cite{lea2016}), mAP@mid is the mean average precision with mid point hit criterion and accuracy denotes the frame wise accuracy. NA indicates that the metric is unavailable in the respective baseline method.}
\label{tab:tab_new}
\vspace{-5mm}
\end{table}
\begin{figure}[!htb]
 \centering
 \includegraphics[width=.95\linewidth]{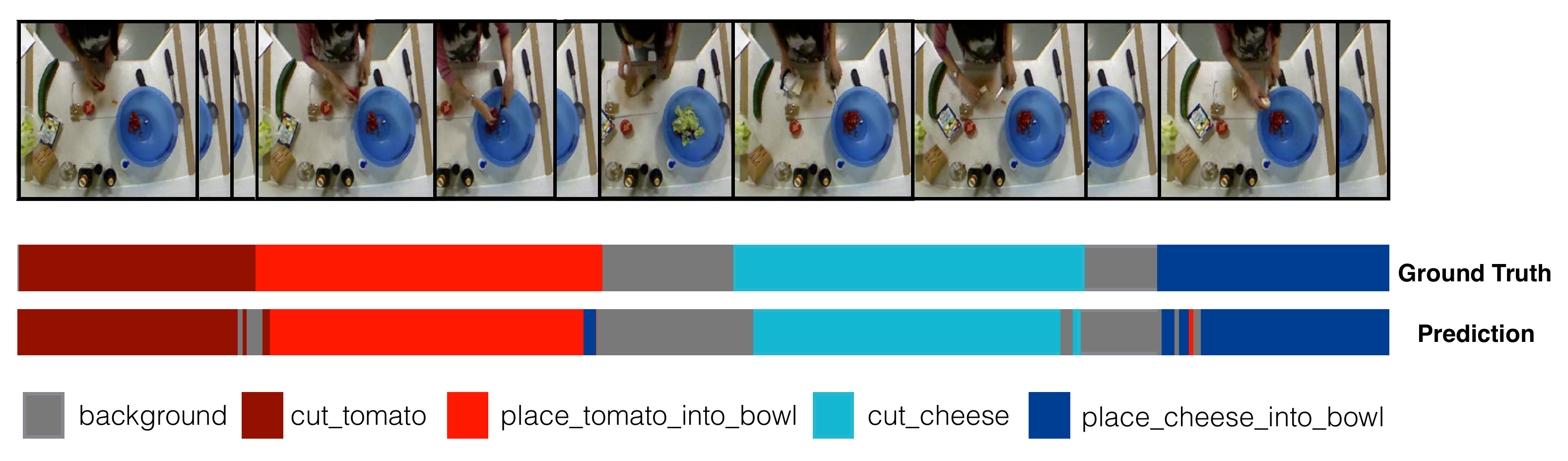} 
 \caption{\textit{Coupled Action GAN} prediction results for the 50 Salads dataset.}
\label{fig:result_1}
\end{figure}

Figure \ref{fig:result_1} shows prediction outputs obtained from the \textit{coupled action GAN} model for the 50 Salads dataset. We observe that there are several areas where the actions have been confused with the $background$ class. For example, the actions such as $cut\_tomato$, $cut\_cheese$, $place\_cheese\_into\_bowl$.  While some classification errors occur, these are typically at the event boundaries where it is difficult to precisely determine action transitions. Overall, we see that all true events are detected and false detected events only last for very short periods of time.

\begin{figure}[!htb]
    \centering
    \begin{subfigure}{.45\linewidth}
        \includegraphics[width=.95\columnwidth]{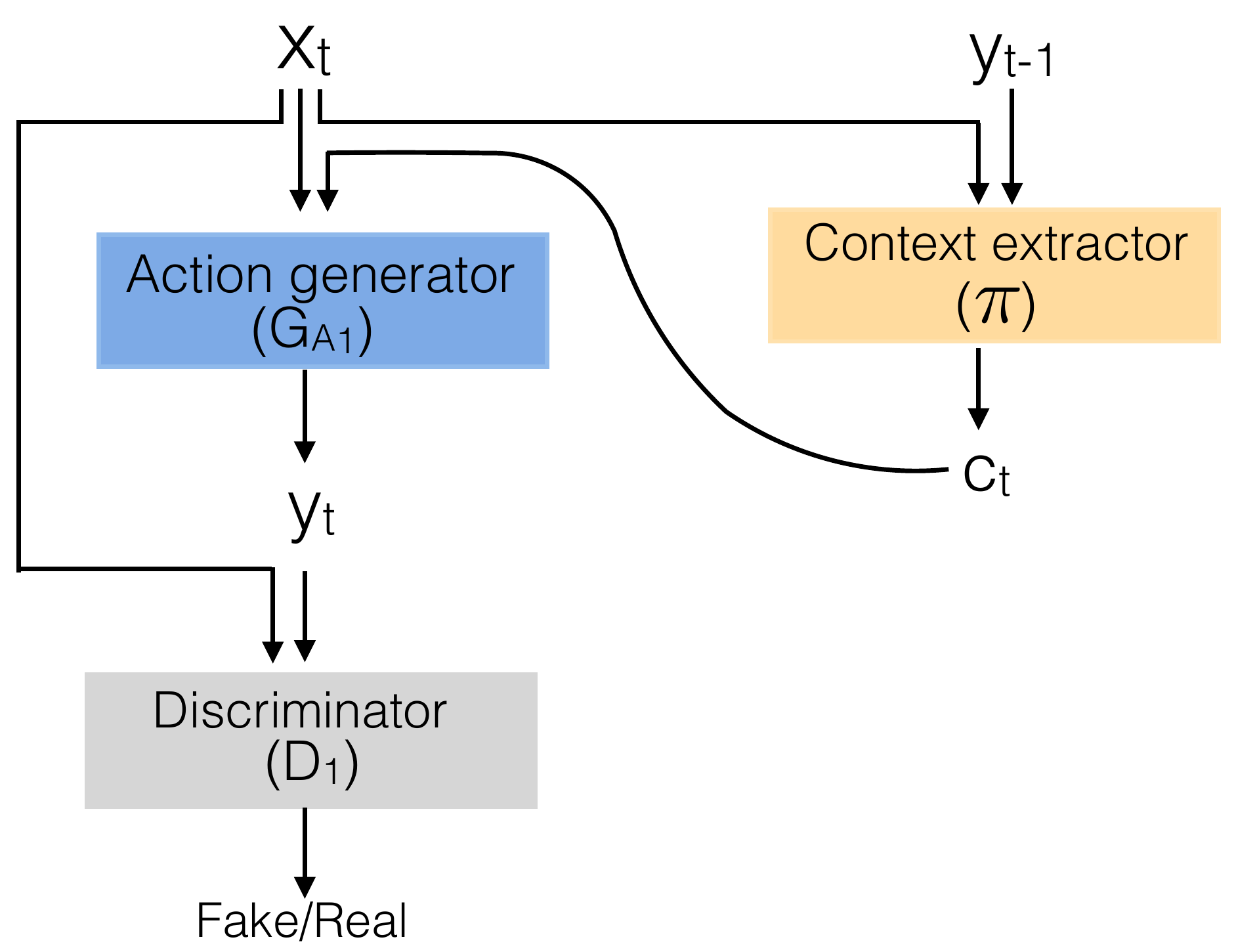} %
        \caption{}
        \label{fig:a}
    \end{subfigure}
    \begin{subfigure}{.2\linewidth}
        \includegraphics[width=.95\columnwidth]{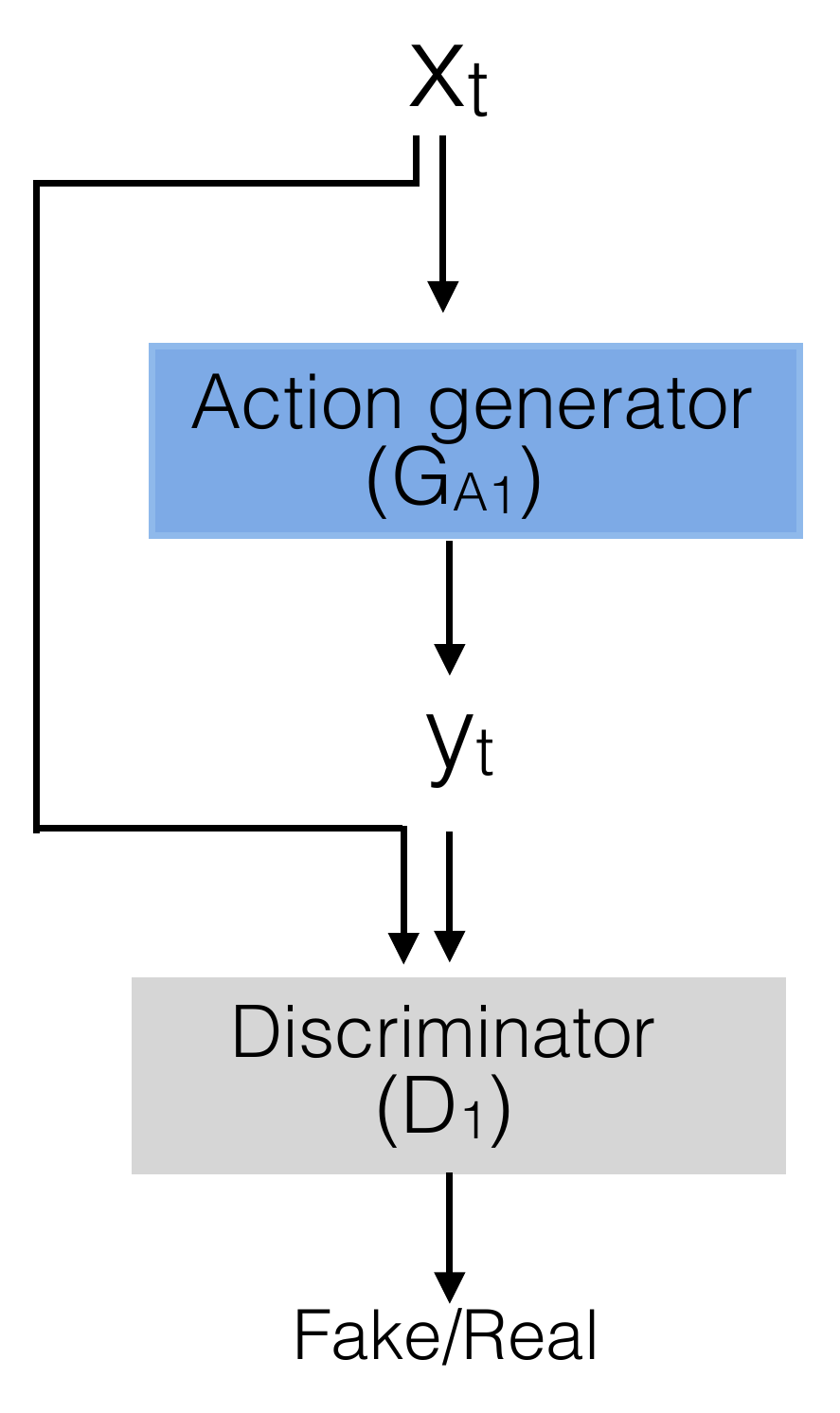}
       \caption{}
       \label{fig:b}
     \end{subfigure}
     \vspace{-2mm}
 \caption{Simplified versions of the \textit{coupled action GAN} model: \textit{unitary GAN} (a) and \textit{unitary GAN - context} (b) }
  \label{fig:act_GAN}
\vspace{-3mm}
\end{figure}

\subsection{Ablation Experiments}
\vspace{-2mm}
\label{ablation}
We evaluate the \textit{coupled action GAN} model against a series of counter parts which are strategically developed by removing components from the proposed model. The details of these baseline models are as follows.
\begin{enumerate}[label=\alph*]

\item (\textit{$G_{A_{1}}$}) : A supervised model created by removing the discriminator and context extractor from the GAN architecture in \ref{fig:b}, and back propagating the classification error.
\item (\textit{$G_{A_{1}}$ + context}): The model of (a) with the context extractor added
\item (\textit{$G_{A_{1}}$ + context + $G_{A_{2}}$}): Couple the model of (b) with $G_{A_{2}}$. Here we jointly back propagate the classification of the two generator networks.
\item (\textit{conditional GAN}) : \textit{unitary GAN} model without the context extractor and objective reinforcement. Hence it is the standard conditional GAN model and follows the objective given in Equation \ref{eq:3}.
\item (\textit{unitary GAN - context}) : \textit{unitary GAN} model without the context extractor, see Figure \ref{fig:b}.
\item (\textit{unitary GAN - $L_c$}) : \textit{unitary GAN} model without classification error based objective reinforcement, optimising the objective given in Equation \ref{eq:4}.
\item (\textit{unitary GAN}) : \textit{coupled action GAN} model without the auxiliary network, optimising the objective given in Equation \ref{eq:6} and depicted in Figure \ref{fig:a}.

\end{enumerate}

Using the same experimental settings as in Section \ref{results}, we present the evaluations for the seven reduced models along with results obtained \textit{coupled action GAN} in Table \ref{tab:tab_5} for the MERL shopping dataset. 

$G_{A_{1}}$ can be considered the simplest model. $G_{A_{1}}$ is not supported by the context extractor, and does not acquire any information regarding the previous time step. Therefore, this naive model simply associates the input frame to an action code without incorporating historic data.

From the results for the model \textit{$G_{A_{1}}$ + context}, it is evident that the context information has the ability improve performance, with the frame wise accuracy increasing by 16\% and the mAP@mid by 17\%. With the context extractor the overall model becomes a recurrent model. In the model \textit{$G_{A_{1}}$ + context + $G_{A_{2}}$}, the additional information stream improves context information, similar to multimodal streams in MSN Det \cite{merlshopping}. However, the model could not achieve the performance of approaches such as MSN Seg \cite{merlshopping}, Dilated TCN \cite{leaCVPR} and ED- TCN \cite{leaCVPR} due to inherent deficiencies \cite{Isola_CVPR2017} when learning with generic loss functions.

We would like to separate the models, \textit{$G_{A_{1}}$}, \textit{$G_{A_{1}}$ + context} and \textit{$G_{A_{1}}$ + context + $G_{A_{2}}$} from the rest of the ablation models. The former are generic supervised models which simply map pixels to action classes and learn this mapping through back-propagating an off the shelf classification loss. However with the introduction of the GAN framework with task specific loss function learning, we achieve a significant performance boost compared to these simpler ablation models and the baseline models (see Table \ref{tab:tab_new}).  With \textit{$G_{A1}$} we perform a softmax classification where we map input pixels to probabilistic labels. This is trivial when the representation and structure of classes are unique, but challenging in continuous action segmentation where background frames appear visually similar to action frames, and action occurrences are related. With the conditional GAN we learn an objective which maps the input to an intermediate representation (i.e action codes) which is easily distinguishable by the classifier. The merit of the intermediate representation is shown by the performance gap between $G_{A1}$ and the conditional GAN. Without sophisticated temporal modelling or very deep feature extraction schemes, even the simplest form of the proposed GAN framework has been able to outperform the baselines by a significant margin. We build upon this observation adding high level context information to the GAN framework and developing a recurrent model to attain sequence modelling. 

\begin{table}[!t]
\begin{center}
\resizebox{0.9\linewidth}{!}{
\begin{tabular}{|c|c|c|c|}
 \hline

        Approach  &F1@\{10,25,50\} & mAP@mid & accuracy \\	
  \hline
  	 \textit{$G_{A1}$}& 30.8, 24.1, 19.2 & 31.3 & 29.7 \\ 
	 \textit{$G_{A1}$ + context}& 49.0, 48.4, 44.2 & 48.3 & 46.1 \\ 
	 \textit{$G_{A1}$ + context + $G_{A2}$}&63.6, 62.2, 55.4 & 61.7  & 58.4 \\ 
 \hline	 
	 \textit{conditional GAN}& 86.9, 85.3, 73.7 & 77.1 & 81.3 \\ 
	 \textit{unitary GAN - context}& 87.2, 85.6, 79.8 & 80.8 & 86.6 \\ 
	 \textit{unitary GAN - $L_c$}&89.0, 86.7, 80.9 & 83.7  & 87.8 \\ 
	 \textit{unitary GAN}& 89.5, 87.3, 81.0 & 84.3  & 88.2 \\ 
	 \textit{coupled action GAN}& \textbf{92.8}, \textbf{91.7}, \textbf{86.2} & \textbf{89.8}  & \textbf{92.6}\\ 
	
\hline
			
\end{tabular} }
\end{center}
\vspace{-5mm}
\caption{Ablation experiment results for MERL Shopping}\label{tab:tab_5}
\vspace{-5mm}
\end{table}

\begin{figure*}[!h]
    \centering
    \begin{subfigure}{.45\linewidth}
             \includegraphics[width=.95\columnwidth]{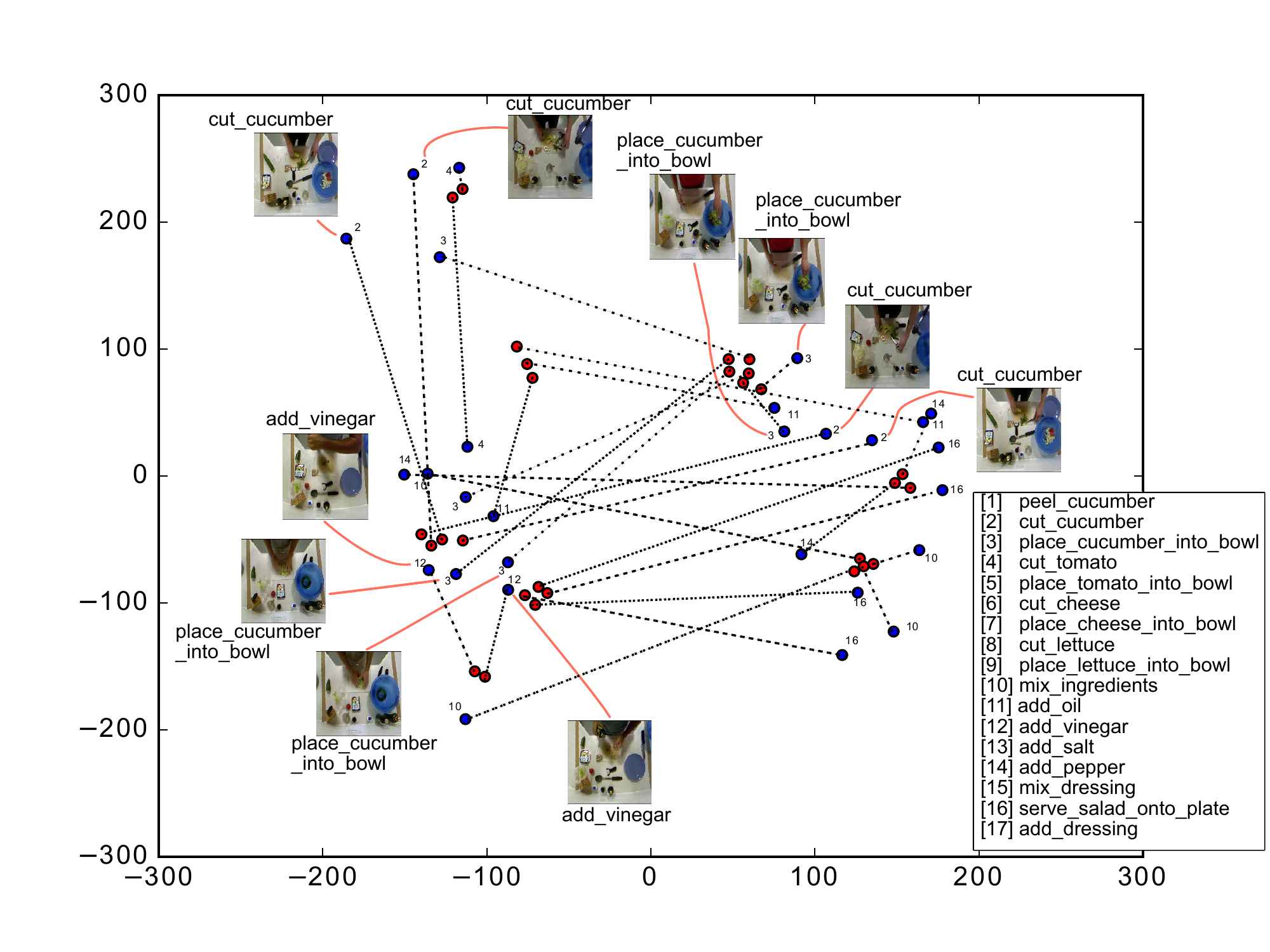} %
        \caption{from $G_{A_1}$ of the proposed \textit{coupled action GAN}}
 	\label{fig:a2}
    \end{subfigure}
    \begin{subfigure}{.45\linewidth}
        \includegraphics[width=.95\columnwidth]{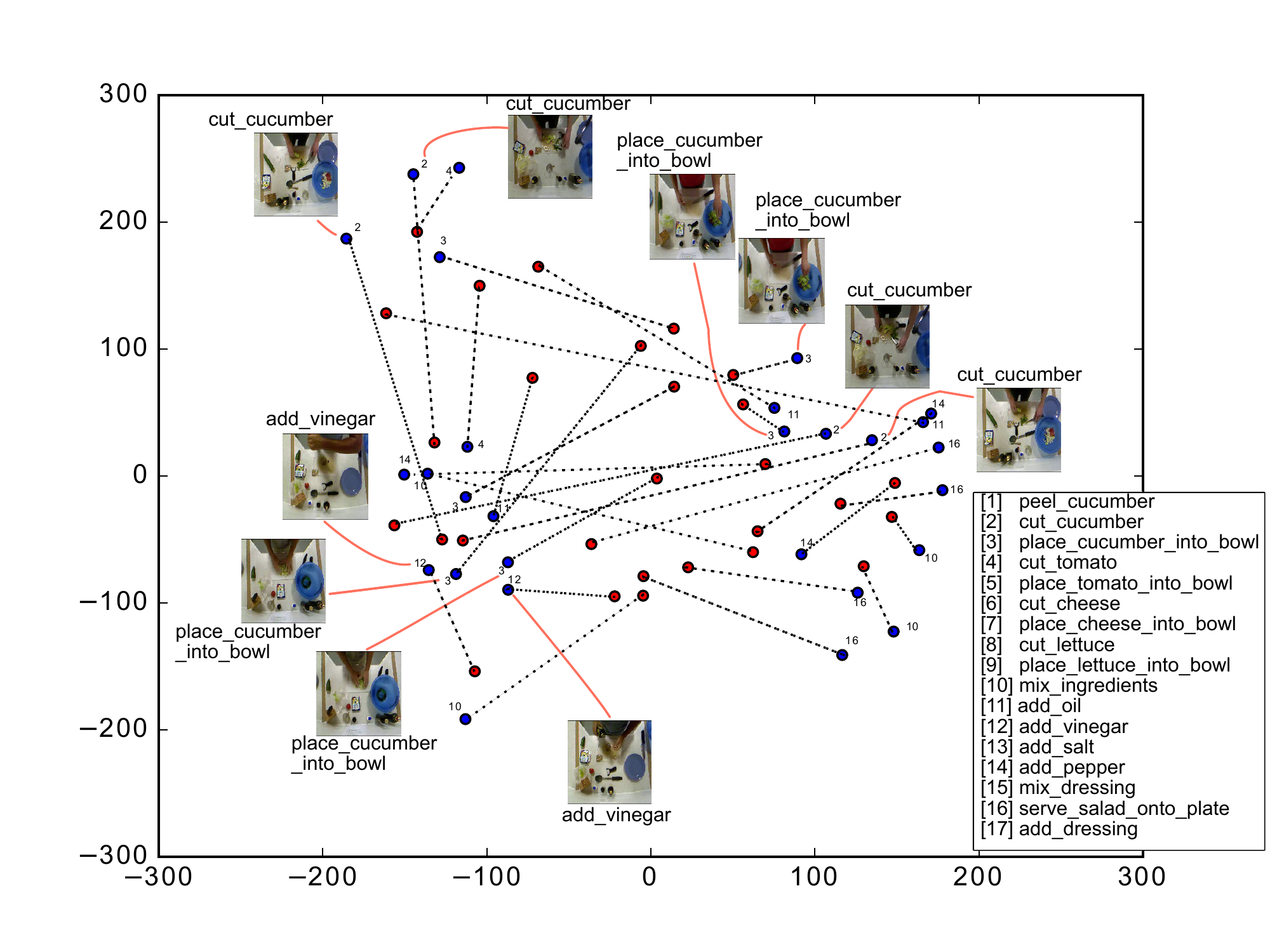}
       \caption{from $G_{A_1}$ of the ablation model $G_{A_1}$ + context + $G_{A_2}$}
      \label{fig:b2}
     \end{subfigure}
     \vspace{-2mm}
 \caption{The visualisations of the embedding space positions before (in blue) and after (in red) the training}
  \label{fig:embed_shift}
\end{figure*}

We observe a gain in performance from \textit{conditional GAN} to \textit{unitary GAN - context} due to the use of classification loss based reinforcement of the GAN objective. Furthermore, comparing the accuracies for \textit{unitary GAN - context} and \textit{unitary GAN - $L_c$}, emphasises the importance of proper context localisation in order to recognise complex human actions. The incorporation of context provides rich feature embeddings to drive the action classifier, hence playing a significant role in the proposed framework. 

Temporal context localisation and classification loss based reinforcement of the objective further boosts performance of the \textit{unitary GAN} model. Yet we observe a significant performance improvement from \textit{unitary GAN} to \textit{coupled action GAN}, demonstrating the importance of auxiliary input streams for capturing fine grain motion and behavioural patterns. Importantly, we also note that all GAN models in Table \ref{tab:tab_5} outperform the state-of-the-art methods in Table \ref{tab:tab_new}, highlighting the benefits that the cGAN architecture offers for continuous fine grained action recognition.

To demonstrate performance if the classifier receives features from both \textit{$G_{A_{1}}$} and \textit{$G_{A_{2}}$}, we conduct a further experiment as follows. Following the approach for $G_{A1}$ (see Section 3.2), we concatenate $G_{A1}$ and $G_{A2}$ features as classifier input. Performance for MERL is 90.1 (accuracy) and  87.3 (mAP@mid), both slightly lower than the proposed method. We believe this is due to the added redundant auxiliary features, and the simple softmax classification layer having insufficient capacity to untangle these redundant features. While more capacity could be added, this would increase complexity. Similarly, when classification uses only ${G_{A2}}$ features; we obtain 89.7 (accuracy) and 86.8 (mAP@mid), indicating the RGB stream is more informative.
     
%
%

\subsection{Effectiveness of Adversarial $+$ Supervised loss combination}
\vspace{-2mm}

To further demonstrate the discriminative ability of the proposed model we have conduct the following experiment. We select 30 examples from the validation set of the 50 Salads dataset. These examples are chosen from the validation set and include different subjects performing different actions. However, appearance wise all examples exhibit similar characteristics with changes between samples primarily being the hand and object positions. Figure \ref{fig:a2} shows the visualisations of the embedding space positions before (in blue) and after (in red) training $G_{A1}$ in the proposed framework. Figure \ref{fig:b2} visualises the same embedding space positions for $G_{A1}$ of the ablation model $G_{A1}$ + context + $G_{A2}$, which is trained using a supervised classification loss. Following \cite{aubakirova2016interpreting} we extracted activations from layer 5 and applied PCA \cite{wold1987principal} to plot them in 2D. The respective ground truth class IDs are indicated in brackets. 

From Figure \ref{fig:a2} it is clear that the frames from the same action class are more tightly grouped by the proposed coupled action GAN, while the supervised learning model is having difficulties learning the common nature of examples within the same action class. This proper grouping leads the introduced model to achieve better classification results. With the GAN learning framework, the generator learns a synthetic objective function that forces it to embed frames from similar action classes closely. This simplifies the task of the action classifier, allowing us to obtain a substantial improvement in performance. The supervised model (Figure \ref{fig:b2}) struggles to obtain a proper grouping as is done by the proposed model, where the supervised model embeddings are only loosely grouped after training.

\subsection{Time Complexity}
\vspace{-2mm}
We evaluate the time consumption of \textit{coupled action GAN} and the model generates 500 predictions in 24.2 seconds using a single core of an Intel E5-2680 2.50GHz CPU. We also evaluate the \textit{unitary GAN} ablation model  (i.e no auxiliary stream) which makes 500 predictions in 14.1 seconds. We used OpenCV toolbox for optical flow computation and it takes 10.4 seconds for 500 frames. Depth information is already available in the 50 Salads \cite{50salads} dataset. 

\section{Conclusion}
\vspace{-2mm}
In this paper, we propose a coupled GAN framework for fine grain human action segmentation in video. The proposed model utilises RGB frames and auxiliary information to better model the evolution of human actions within the given video sequence, outperforming state-of-the-art methods on three datasets: the 50 Salads, MERL Shopping and Georgia Tech Egocentric Activities dataset. Evaluations on both static and dynamic cameras with overhead and egocentric view demonstrate the importance of the architectural augmentations proposed in this framework for segmenting fine grain actions. We show the highly beneficial nature of capturing auxiliary information, not only to boost performance but also to ensure the flexibility of the system to adapt to different information cues provided in different datasets. Even through we perform evaluation on unsegmented continuous action datasets, which is comparatively more challenging, the proposed system can be directly applied to pre-segmented datasets for action recognition.

{\small
\bibliographystyle{ieee}
\bibliography{paper3}
}

\end{document}